%% file: paper.tex
\documentclass[ijoc]{informs_arxiv}

\usepackage{amsmath}
\usepackage{amssymb}
\usepackage{amsfonts}
\usepackage{booktabs}
\usepackage{multirow}
\usepackage{graphicx}
\usepackage{subfig}
\usepackage{siunitx}

\usepackage{natbib}
\bibpunct[, ]{(}{)}{,}{a}{}{,}%
\def\bibfont{\small}%

\OneAndAHalfSpacedXI
\TheoremsNumberedThrough
\ECRepeatTheorems
\EquationsNumberedThrough

\JOURNAL{INFORMS Journal on Computing}
\MANUSCRIPTNO{}

\DeclareMathOperator*{\maximize}{\mathrm{maximize}}
\newcommand{\laplacian}{\nabla^2}
\newcommand{\ldiff}[2]{\frac{d #1}{d #2}}
\newcommand{\ldiffn}[3]{\frac{d^{#3} #1}{d #2^{#3}}}
\newcommand{\lpdiff}[2]{\frac{\partial #1}{\partial #2}}
\newcommand{\lpdiffn}[3]{\frac{\partial^{#3}\!#1}{\partial #2^{#3}}}
\newcommand{\R}{\mathbb{R}}

\newcommand{\sF}{\mathcal{F}}
\newcommand{\sN}{\mathcal{N}}
\newcommand{\sP}{\mathcal{P}}
\newcommand{\bdA}{\mathbf{A}}
\newcommand{\bdB}{\mathbf{B}}
\newcommand{\bdC}{\mathbf{C}}
\newcommand{\bdD}{\mathbf{D}}
\newcommand{\bdF}{\mathbf{F}}
\newcommand{\bdI}{\mathbf{I}}
\newcommand{\bdP}{\mathbf{P}}
\newcommand{\bdPNFtilde}{\bdP_{\sN(\tilde{\bdF})}}
\newcommand{\bdQ}{\mathbf{Q}}
\newcommand{\bdK}{\mathbf{K}}
\newcommand{\bdM}{\mathbf{M}}
\newcommand{\bdS}{\mathbf{S}}
\newcommand{\bdY}{\mathbf{Y}}
\newcommand{\bdd}{\mathbf{d}}
\newcommand{\bdf}{\mathbf{f}}
\newcommand{\bdg}{\mathbf{g}}
\newcommand{\bdq}{\mathbf{q}}
\newcommand{\bds}{\mathbf{s}}
\newcommand{\bdu}{\mathbf{u}}
\newcommand{\bdy}{\mathbf{y}}
\newcommand{\bddelta}{\boldsymbol{\delta}}
\newcommand{\bdlambda}{\boldsymbol{\lambda}}
\newcommand{\ttt}[1]{\times 10^{#1}}

\let\Re\undefined
\DeclareMathOperator{\Re}{\mathrm{Re}}
\let\Im\undefined
\DeclareMathOperator{\Im}{\mathrm{Im}}

\begin{document}

\TITLE{Adversarial Perturbations of Physical Signals}
\RUNTITLE{Adversarial Perturbations of Physical Signals}

\ARTICLEAUTHORS{%
\AUTHOR{Robert L.\ Bassett, Austin Van Dellen}
\AFF{Department of Operations Research, Naval Postgraduate School, 1 University Cir., Monterey, CA 93943, USA \EMAIL{robert.bassett@nps.edu}, \EMAIL{austin.vandellen@nps.edu}}
\AUTHOR{Anthony P. Austin}
\AFF{Department of Applied Mathematics, Naval Postgraduate School, 1 University Cir., Monterey, CA 93943, USA \EMAIL{anthony.austin@nps.edu}}
}
\RUNAUTHOR{Bassett, Van Dellen, and Austin}

\ABSTRACT{%
We investigate the vulnerability of computer-vision-based signal classifiers to
adversarial perturbations of their inputs, where the signals and perturbations
are subject to physical constraints.  We consider a scenario in which a source
and interferer emit signals that propagate as waves to a detector, which
attempts to classify the source by analyzing the spectrogram of the signal it
receives using a pre-trained neural network.  By solving PDE-constrained
optimization problems, we construct interfering signals that cause the detector
to misclassify the source even though the perturbations to the spectrogram of
the received signal are nearly imperceptible.  Though such problems can have
millions of decision variables, we introduce methods to solve them efficiently.
Our experiments demonstrate that one can compute effective and physically
realizable adversarial perturbations for a variety of machine learning models
under various physical conditions.
}

\FUNDING{All authors acknowledge support from ONR grants N0001421WX00142 and
N0001423WX01316.}

\KEYWORDS{Adversarial perturbations, machine learning, neural networks, PDE-constrained optimization}

\maketitle

\section{Introduction}
\label{SEC:Introduction}
\input{introduction.tex}

\section{Problem Formulation}
\label{SEC:Formulation}
\input{formulation.tex}

\section{Discretization}
\label{SEC:Discretization}
\input{discretization.tex}

\section{Computation}
\label{SEC:Computation}
\input{computation.tex}

\section{Experimental Results}
\label{SEC:Experiments}
\input{experiments.tex}

\section{Conclusion}
\label{SEC:Conclusion}
\input{conclusion.tex}

\ACKNOWLEDGMENT{We thank the High Performance Computing Center at the Naval
Postgraduate School for supplying the computing resources used for our
experiments.

The views expressed in this document are those of the authors
and do not reflect the official policy or position of the Department of
Defense, the Department of the Navy, the Office of Naval Research, or the
U.~S.\ Government.}

\bibliographystyle{informs2014}
\bibliography{thebib}

\end{document}

%% file: introduction.tex
Advances in machine learning have produced advances in signal classification
with perhaps the greatest leaps occurring in the field of computer vision.
Many methods have been proposed that leverage techniques from computer vision
to classify non-visual signals by representing them as images.  But despite
their successes, computer vision methods have been shown to be susceptible to
\emph{adversarial perturbations}, minor modifications to an input image that
cause dramatic changes in a classifier's output.  The existence of adversarial
perturbations demonstrates that many types of classifiers---in particular, deep
neural networks---are not robust to small changes in their inputs.

In the wake of the seminal work of \cite{SZSBEGF2014}, many methods have been
developed for constructing adversarial perturbations for neural network
classifiers. Notable foundational examples include the fast gradient sign
method of \cite{GSS2015}, Deepfool \citep{MFF2016}, and Carlini--Wagner
perturbations \citep{CW2017}.  The zeroth-order optimization approach of
\cite{CZSYH2017} provides a way to compute adversarial perturbations in the
absence of gradient information.  \cite{BGR2020} construct perturbations which
are less perceptible to an observer by accounting for human perception of color
and texture.

All of these methods construct perturbations under the assumption that the
adversary has what we call \emph{logical access} to the classifier, i.e., the
ability to modify inputs to the classifier directly. For computer vision,
logical access is often an inappropriate assumption because in many
applications, the classifier receives its input from a sensor, such as a
camera, and an adversary has only \emph{physical access} to the classifier via
the sensor.  That is, the adversary can only influence the inputs to the
classifier by manipulating the sensor's physical environment.  Physical access
is a weaker assumption on an adversary's capabilities than logical access
because constructing a perturbation that can be physically realized is more
difficult than constructing a perturbation using digital artifacts that cannot
be replicated in the sensor's environment.

Computer vision researchers have successfully constructed adversarial
perturbations under physical access assumptions \citep{KGB2018,SBBR2016}.  In
these works, the inputs to the neural networks are direct images of the
physical objects being classified.  In other applications of computer vision to
signal classification, the relationship between the objects and their digital
representations is less direct.  Here, we consider applications in which
univariate signals to be classified are represented as \emph{spectrograms},
images that show how the signals' frequency content changes with time
\citep[Chapter 7]{Coh1995}.  Classifying signals via their spectrograms is both
simple (since one can leverage off-the-shelf techniques from computer vision)
and effective, having come to define the state of the art in several areas
\citep{liu2023asvspoof, sharma2020trends}.

Motivated by the success and popularity of spectrogram classification
techniques, we investigate their susceptibility to adversarial perturbations
under physical access assumptions.  We model the physical access constraints
using partial differential equations (PDEs), leading to PDE-constrained
optimization problems for the perturbations.  These problems are considerably
more complicated than the optimization problems typically encountered in the
adversarial perturbations literature; we show how to solve them efficiently.
We also examine the effects of environmental noise, which usually has a larger
impact on the kinds of signals classified by spectrogram---such as acoustic or
(non-optical) electromagnetic signals---than it does for images.

%% file: formulation.tex
To make our discussion concrete, we consider the following scenario.  An
intruding submarine wishes to pass unnoticed through a region of seawater
containing an acoustic detector.  To accomplish this, the submarine employs an
external interferer, which emits a perturbing signal to confuse the detector.
The detector operates by using a neural network to process spectrograms of the
acoustic signals it receives.  Our aim is to select a perturbing signal that
will cause the detector to misclassify a spectrogram and conclude that the
submarine is not present, even though the perturbed spectrogram differs little
from the unperturbed one.

\begin{figure}[t]
\centering
\includegraphics{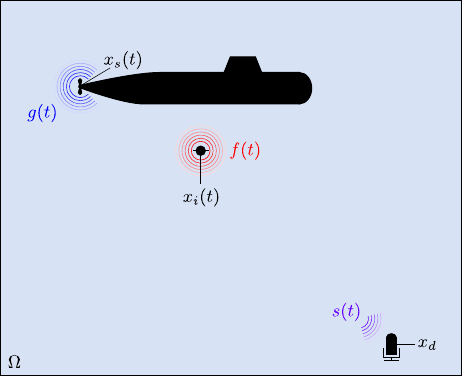}
\caption{Problem setup.  The submarine at $x_s(t)$ emits an acoustic signal
$g(t)$.  The interferer at $x_i(t)$ emits a perturbing signal $f(t)$.  The
detector at $x_d$ receives a signal $s(t)$, the result of $f$ and $g$
propagating through $\Omega$ as waves.}
\label{FIG:Setup}
\end{figure}

Figure \ref{FIG:Setup} illustrates the general setup.  Our region of water is
is a rectangle $\Omega \subset \R^2$.  The submarine and interferer are located
at positions $x_s$ and $x_i$, respectively, which may vary with time.  The detector
is located at a fixed position $x_d$.  As it moves, the submarine emits a
signal $g(t)$.  We want to choose the signal $f(t)$ emitted by the interferer
to fool the detector.

We model the propagation of sound in $\Omega$ using the linear wave equation
\citep[Section 5.5]{KFCS2000},
\begin{equation}\label{EQN:WaveEqn}
\lpdiffn{u}{t}{2} = c^2 \laplacian u + q(x, t),
\end{equation}
where $u$ is the acoustic pressure, $c$ is the speed of sound in water, and
$\laplacian$ is the Laplacian.  The forcing term $q(x, t)$ contains the signals
generated by the submarine and the interferer, which we model as point sources,
taking $q$ to have the form
\[
q(x, t) = f(t)\delta_\varepsilon\bigl(x - x_i(t)\bigr) + g(t)\delta_\varepsilon\bigl(x - x_s(t)\bigr),
\]
where $\varepsilon$ is a small parameter and
\[
\delta_\varepsilon(x) = \frac{\varepsilon^2}{\pi^2(x_1^2 + \varepsilon^2)(x_2^2 + \varepsilon^2)}
\]
approximates the Dirac delta \citep[Section 2.5]{GS1964}.  For simplicity, we
assume that $\Omega$ is initially devoid of sound, yielding the zero initial
conditions
\[
u(x, 0) = 0 \quad\text{and}\quad \lpdiff{u}{t}(x, 0) = 0.
\]
Additionally, we take $\Omega$ to be a region of open water away from the
seafloor and surface so that the effects of these boundaries may be neglected.
To model the propagation of waves out through the boundary $\partial \Omega$ of
$\Omega$, we use a first-order approximation to an absorbing boundary condition
\citep{EM1977} \citep[Section 6.1.2]{CP2017},
\begin{equation}\label{EQN:AbsorbingBC}
\lpdiff{u}{t} + c \nabla u \cdot \hat{n} = 0, \quad x \in \partial \Omega,
\end{equation}
where $\hat{n}$ denotes the outward-pointing unit normal vector to $\partial
\Omega$.

The detector computes a spectrogram $\hat{s} = 10 \log_{10} |\sF s|^2$ in
decibels (dB), of the signal $s(t) = u(x_d, t)$ that it receives, where $\sF$
is the short-time Fourier transform operator and $|\cdot|$ denotes the
elementwise modulus of a complex function.  The detector then passes $\hat{s}$
through a neural network $\Phi$ that has been trained to recognize spectrograms
produced by the submarine when the interferer is absent.  Let $L$ denote the
loss function used to train the network; this is a function that takes
$\Phi(\hat{s})$ as one argument, together with a label $y$ that assumes the
value $1$ (respectively, $0$) when the submarine is present (respectively,
absent).  To fool the detector, we want to select $f$ to make
$L\bigl(\Phi(\hat{s}), y\bigr)$ as large as possible, and to model the
requirement that $f$ be emittable by a real speaker, we constrain its frequency
content to lie in a predetermined band.  That is, we demand
\[
\sP \sF f = 0,
\]
where $\sP$ projects onto the frequencies that are disallowed.

We have thus arrived  at the following optimization problem for the perturbing
signal $f$:
\begin{equation}\label{EQN:OptProblem}
\begingroup
\renewcommand{\arraystretch}{1.25}
\begin{tabular}{ll}
$\displaystyle\maximize_{f, u, s, \hat{s}}$ & $J(f) = L\bigl(\Phi(\hat{s}), y\bigr)$ \\
\text{subject to} & $\lpdiffn{u}{t}{2} = c^2 \laplacian u + f(t)\delta_\varepsilon\bigl(x - x_i(t)\bigr) + g(t)\delta_\varepsilon\bigl(x - x_s(t)\bigr)$ \\
& $u(x, 0) = \lpdiff{u}{t}(x, 0) = 0$ \\
& $\lpdiff{u}{t} + c \nabla u \cdot \hat{n} = 0, \quad x \in \partial \Omega$ \\
& $s(t) = u(x_d, t)$ \\
& $\hat{s} = 10 \log_{10} |\sF s|^2$ \\
& $\sP \sF f = 0.$
\end{tabular}
\endgroup
\end{equation}
What makes this optimization problem different from those usually encountered
in the literature on adversarial perturbations is the nature of the constraint.
In our scenario, the perturbing signal cannot interact with the input of the
neural network directly; the interaction is mediated by the physics of the
surrounding environment as described by the PDE \eqref{EQN:WaveEqn}.

%% file: discretization.tex
To discretize \eqref{EQN:WaveEqn} in space, we use a continuous Galerkin finite
element method \citep{Hug2000}.  Applying standard manipulations, we arrive at
the variational formulation
\[
\ldiffn{}{t}{2} \int_\Omega u v \: dV = -c^2 \int_\Omega \nabla u \cdot \nabla v \: dV - c \ldiff{}{t} \int_{\partial \Omega} u v \: dS + \int_\Omega q v \: dV
\]
for \eqref{EQN:WaveEqn} with the boundary condition \eqref{EQN:AbsorbingBC},
where $v$ is a test function.  We partition $\Omega$ into a mesh of conforming,
nonoverlapping triangular elements and approximate $u$ and $q$ on the mesh by
continuous piecewise linear interpolants over the elements.  At each time $t$,
the interpolants are determined by the values of $u(x, t)$ and $q(x, t)$ at the
mesh nodes (a $P_1$ Lagrange representation), which we gather into column
vectors $\bdu(t)$ and $\bdq(t)$.  Imposing the Galerkin condition yields the
system of ODEs
\begin{equation}\label{EQN:SemidiscreteSystem}
\bdM \bdu'' = -c^2 \bdK \bdu - c \bdS \bdu' + \bdM \bdq
\end{equation}
for $\bdu$, where $\bdM$, $\bdS$, and $\bdK$ are the standard finite element
mass, surface mass, and stiffness matrices.

We discretize \eqref{EQN:SemidiscreteSystem} using a leapfrog scheme
\citep[Section 5.3]{LeV2007}.  For a single ODE $u'' = f(u, t)$, the update
equation for the scheme is
\begin{equation}\label{EQN:Leapfrog}
\frac{u^{k + 1} - 2u^k + u^{k - 1}}{(\Delta t)^2} = f(u^k, t_k),
\end{equation}
where $\Delta t$ is a chosen time step and $u^k$ is the approximation to the
solution at time $t_k = k(\Delta t)$, $k = 0, \ldots, K$.  This is an explicit
scheme, second-order accurate, and is nondissipative, making it well-suited to
undamped wave propagation problems, provided the wave numbers are low enough to
avoid dispersive artifacts \citep[Section 10.2.2]{LeV2007}.  Applying
\eqref{EQN:Leapfrog} to \eqref{EQN:SemidiscreteSystem} and rearranging terms,
we obtain
\begin{equation}\label{EQN:DiscreteSystem}
\left(\bdM + \frac{c(\Delta t)}{2} \bdS\right)\bdu^{k + 1} = \left(2\bdM - c^2 (\Delta t)^2 \bdK\right)\bdu^k + \left(\frac{c(\Delta t)}{2} \bdS - \bdM\right) \bdu^{k - 1} + (\Delta t)^2 \bdM \bdq^k,
\end{equation}
where $\bdq^k = \bdq(t_k)$ and $\bdu^k$ approximates $\bdu(t_k)$.

Equation \eqref{EQN:DiscreteSystem} defines a sequence of $K - 1$ linear
systems that can be solved to find $\bdu^2, \ldots, \bdu^K$ given initial data
$\bdu^0$ and $\bdu^1$.  To facilitate further discussion, it is useful to
express these as a single system for all the $\bdu^k$ at once.  (This is just
for notational convenience; we do not form this system in our implementation,
which uses \eqref{EQN:DiscreteSystem} directly.)  Abusing notation, we redefine
$\bdu$ to be the block column vector
\[
\bdu = \begin{bmatrix}
\bdu^0 \\ \vdots \\ \bdu^K
\end{bmatrix}
\]
and introduce the the block upper-triangular matrix
\[
\newcommand{\blkI}{\bdI}
\newcommand{\blkAm}{\bdA^-}
\newcommand{\blkAz}{\bdA^0}
\newcommand{\blkAp}{\bdA^+}
\bdA = \begin{bmatrix}
\blkI  &        &        &        &        & \\
       & \blkI  &        &        &        & \\
\blkAm & \blkAz & \blkAp &        &        & \\
       & \blkAm & \blkAz & \blkAp &        & \\
       &        & \ddots & \ddots & \ddots & \\
       &        &        & \blkAm & \blkAz & \blkAp
\end{bmatrix},
\]
where $\bdI$ is the identity matrix and
\[
\bdA^- = \bdM - \frac{c(\Delta t)}{2}\bdS, \qquad
\bdA^0 = c^2 (\Delta t) ^{2} \bdK - 2\bdM, \qquad
\bdA^+ = \bdM + \frac{c(\Delta t)}{2}\bdS.
\]
Further, we decompose the forcing vector $\bdq^k$ as
\[
\bdq^k = f(t_k) \bddelta_i^k + g(t_k) \bddelta_s^k,
\]
where $\bddelta_i^k$ and $\bddelta_s^k$ contain the values of
$\delta_\varepsilon\bigl(x - x_i(t_k)\bigr)$ and $\delta_\varepsilon\bigl(x -
x_s(t_k)\bigr)$ at the mesh nodes, respectively, and define the matrices
\[
\bdB = (\Delta t)^2 \begin{bmatrix}
0      &        &                   &                   &        &   \\
       & 0      &                   &                   &        &   \\
       &        & \bdM \bddelta_i^2 &                   &        &   \\
       &        &                   & \bdM \bddelta_i^3 &        &   \\
       &        &                   &                   & \ddots &   \\
       &        &                   &                   &        & \bdM \bddelta_i^{K - 1}
\end{bmatrix},
\quad
\bdC = (\Delta t)^2 \begin{bmatrix}
0      &        &                   &                   &        &   \\
       & 0      &                   &                   &        &   \\
       &        & \bdM \bddelta_s^2 &                   &        &   \\
       &        &                   & \bdM \bddelta_s^3 &        &   \\
       &        &                   &                   & \ddots &   \\
       &        &                   &                   &        & \bdM \bddelta_s^{K - 1}
\end{bmatrix}.
\]
Gathering the values $f(t_0), \ldots, f(t_K)$ and $g(t_0), \ldots, g(t_K)$ into
column vectors $\bdf$ and $\bdg$, respectively, we combine the equations
\eqref{EQN:DiscreteSystem} for $k = 1, \ldots, K - 1$ into a single block
matrix equation,
\[
\bdA \bdu = \bdB \bdf + \bdC \bdg.
\]

%

We have thus completely discretized the PDE constraint in
\eqref{EQN:OptProblem}.  Discretization of the remaining variables and
constraints is more straightforward.  First, we introduce a row vector $\bdd^T$
such that $\bdd^T \bdu^k$ extracts the value at the point $x = x_d$ of the
finite element solution represented by $\bdu^k$.  Defining
\[
\bdD^T = \left. \begin{bmatrix}
\bdd^T \\
& \bdd^T \\
& & \ddots  \\
& & & \bdd^T
\end{bmatrix} \right\} \text{$K + 1$ copies},
\]
we have that $\bds = \bdD^T \bdu$ contains the acoustic pressure observed by the
detector at each time step.  If $\bdF$ is the matrix representing the
short-time Fourier transform (see Section \ref{SSEC:Spectrograms} below), then
the spectrogram of the signal (in dB) at the detector is $\hat{\bds} = 10
\log_{10} |\bdF \bds|^2$.  Finally, let $\bdP$ be a matrix that selects entries
of a spectrogram corresponding to frequencies outside the band of allowable
frequencies for $\bdf$.

Assembling all of these objects, we are led to the following discrete version
of \eqref{EQN:OptProblem}:
\begin{equation}\label{EQN:DiscreteOptProblem}
\begingroup
\renewcommand{\arraystretch}{1.25}
\begin{tabular}{ll}
$\displaystyle\maximize_{\bdf, \bdu, \bds, \hat{\bds}}$ & $J(\bdf) = L\bigl(\Phi(\hat{\bds}), y\bigr)$ \\
\text{subject to} & $\bdA \bdu = \bdB \bdf + \bdC \bdg$ \\
& $\bds = \bdD^T \bdu$ \\
& $\hat{\bds} = 10 \log_{10} |\bdF \bds|^2$ \\
& $\bdP \bdF \bdf = 0.$
\end{tabular}
\endgroup
\end{equation}

%% file: computation.tex
Our goal is to compute local maximizers of \eqref{EQN:DiscreteOptProblem}.  To
do this, we use a first-order descent method with a projection to enforce the
constraint $\bdP \bdF \bdf = 0$ \citep[Section 2.D]{RW2022}.  The primary
computational difficulty is in the evaluations of the objective $J$ and its
gradient, both of which appear to require solution of the constraining PDE.  In
this section, we discuss how to perform these evaluations and the projection
step efficiently.

\subsection{Efficient Evaluation of the Objective}
\label{SUBSEC:shortcut}

Naively, it seems that each evaluation of $J(\bdf)$ requires a solution of the
constraining PDE.  Having selected $\bdf$, one first solves $\bdA \bdu = \bdB
\bdf + \bdC \bdg$ for $\bdu$ using the iteration \eqref{EQN:DiscreteSystem}.
Then, one evaluates $\bds = \bdD^T \bdu$, then $\hat{\bds} = 10 \log_{10} |\bdF
\bds|^2$, and finally $J(\bdf) = L\bigl(\Phi(\hat{\bds}), y\bigr)$.  The solve
for $\bdu$ is by far the most expensive part of this computation, as
\eqref{EQN:DiscreteSystem} entails $K - 1$ solves with the large matrix
$\bdA^+$, which has dimension equal to the number of nodes in the finite
element mesh.  One can reduce the cost by computing and storing a factorization
of $\bdA^+$, but the cost typically remains significant, especially if the mesh
is very fine or the number of time steps is very large.

Remarkably, we can exploit linear algebra to eliminate the need to solve for
$\bdu$ entirely.  The crucial observation is that we are interested not in
$\bdu$ itself but only in
\[
\bds = \bdD^T \bdu = \bdD^T \bdA^{-1}(\bdB \bdf + \bdC \bdg)
\]
and that there are two ways to evaluate this product.  One way evaluates
$\bdA^{-1} (\bdB \bdf + \bdC \bdg)$ first and then multiplies the result by
$\bdD^T$ on the left; this corresponds to the naive approach just described and
requires a solve with $\bdA$ (via \eqref{EQN:DiscreteSystem}) for each value of
$\bdf$.  The other way evaluates $\bdY^T = \bdD^T \bdA^{-1}$ by solving the
adjoint equation
\begin{equation}\label{EQN:YEqn}
\bdA^T \bdY = \bdD
\end{equation}
and then computes
\begin{equation}\label{EQN:sEqn}
\bds = \bdY^T (\bdB \bdf + \bdC \bdg).
\end{equation}
This latter approach has a potential advantage:  because $\bdY$ does not depend
on $\bdf$, we can compute it (or even $\bdY^T \bdB$ and $\bdY^T \bdC$) once,
offline, and then use it to find $\bds$ for as many values of $\bdf$ as needed
in our optimization algorithm.

For this advantage to be realized, the effort required to solve
\eqref{EQN:YEqn} must be less than that required to solve all of the systems
that would arise under the naive approach.  Since $\bdD$ has $K + 1$ columns,
it would seem that solving \eqref{EQN:YEqn} requires $K + 1$ solves with
$\bdA^T$ and, hence, that the alternative approach would be advantageous only
if the number of optimization steps exceeds $K + 1$, under the reasonable
assumption that solves with $\bdA$ and $\bdA^T$ are of similar cost.  In fact,
the situation is considerably better than this: due to the structure of $\bdA$
and $\bdD$, we can solve \eqref{EQN:YEqn} for a cost comparable to that of just
a \emph{single} solve with $\bdA^T$.


In more detail, partitioning $\bdA$ as
\[
\newcommand{\blkI}{\bdI}
\newcommand{\blkAtilde}{\tilde{\bdA}}
\newcommand{\blkAhat}{\hat{\bdA}}
\newcommand{\blkAm}{\bdA^-}
\newcommand{\blkAz}{\bdA^0}
\newcommand{\blkAp}{\bdA^+}
\newcommand{\topstrut}{\rule{0pt}{0.9\normalbaselineskip}}
\newcommand{\botstrut}{\rule[-0.3\normalbaselineskip]{0pt}{0pt}}
\bdA = \left[\begin{array}{c | c}
\blkI & \botstrut \\
\hline
\blkAtilde & \blkAhat\topstrut
\end{array}\right] = \left[\begin{array}{c c | c c c c}
\blkI  &        &        &        &        & \\
       & \blkI  &        &        &        &\botstrut \\
\hline
\blkAm & \blkAz & \blkAp &        &        &\topstrut \\
       & \blkAm & \blkAz & \blkAp &        & \\
       &        & \ddots & \ddots & \ddots & \\
       &        &        & \blkAm & \blkAz & \blkAp
\end{array}\right]
\]
and partitioning $\bdD$ as
\[
\newcommand{\topstrut}{\rule{0pt}{0.9\normalbaselineskip}}
\newcommand{\botstrut}{\rule[-0.3\normalbaselineskip]{0pt}{0pt}}
\bdD = \left[\begin{array}{c | c}
\tilde{\bdD} & \botstrut \\
\hline
             & \hat{\bdD}\topstrut
\end{array}\right] = \left[\begin{array}{c c | c c c c}
\bdd  &        &        &        &        & \\
      & \bdd   &        &        &        &\botstrut \\
\hline
      &        & \bdd   &        &        &\topstrut \\
      &        &        & \bdd   &        & \\
      &        &        &        & \ddots & \\
      &        &        &        &        & \bdd
\end{array}\right],
\]
we have that
\[
\newcommand{\topstrut}{\rule{0pt}{0.9\normalbaselineskip}}
\newcommand{\botstrut}{\rule[-0.3\normalbaselineskip]{0pt}{0pt}}
\bdY = \bdA^{-T}\bdD = \left[\begin{array}{c | c}
\tilde{\bdD} & -\tilde{\bdA}^T \hat{\bdA}^{-T} \hat{\bdD}\botstrut \\
\hline
             & \hat{\bdA}^{-T} \hat{\bdD}\topstrut
\end{array}\right].
\]
The first step in computing $\bdY$ is to solve $\hat{\bdA}^T \hat{\bdY} =
\hat{\bdD}$ for $\hat{\bdY} = \hat{\bdA}^{-T} \hat{\bdD}$.  The matrix
$\hat{\bdA}^T$ is a block upper triangular, block Toeplitz matrix with
invertible diagonal blocks.  The set of such matrices (for a given
partitioning) forms a group under matrix multiplication (see, e.g.,
\citet[Section 0.9.7]{HJ2012} for the non-block case); hence, $\hat{\bdA}^{-T}$
has the same structure.  The matrix $\hat{\bdD}$ is also block upper triangular
(in fact, block diagonal) and block Toeplitz.  The diagonal blocks of
$\hat{\bdD}$ are not invertible---indeed, they are not even square---but the
partitioning of $\hat{\bdD}$ conforms to the partitioning of $\hat{\bdA}^T$
and, hence, to that of $\hat{\bdA}^{-T}$.  It follows that $\hat{\bdY}$, too,
is block upper triangular and block Toeplitz, being a product of two
conformally-partitioned matrices with that structure:
\[
\hat{\bdY} = \begin{bmatrix}
\hat{\bdy}_1 & \hat{\bdy}_2 & \hat{\bdy}_3 & \cdots       & \hat{\bdy}_{K - 1} \\
             & \hat{\bdy}_1 & \hat{\bdy}_2 & \cdots       & \hat{\bdy}_{K - 2} \\
             &              & \ddots       & \ddots       & \vdots       \\
             &              &              & \hat{\bdy}_1 & \hat{\bdy}_2 \\
             &              &              &              & \hat{\bdy}_1 \\
\end{bmatrix}.
\]
We can therefore find all of $\hat{\bdY}$ by finding just its last (block)
column, and that can be accomplished by solving just one system with
$\hat{\bdA}^T$ that has the last column of $\hat{\bdD}$ as the right-hand side.
With $\hat{\bdY}$ in hand, the remaining entries of $\bdY$ can be found
by computing $-\tilde{\bdA}^T \hat{\bdY}$, which is relatively inexpensive.

This reduction in the cost of computing $\bdY$ due to the structure of $\bdA$
and $\bdD$ is more than enough to make the alternative approach to evaluating
$J(\bdf)$ competitive.  By using it, we were able to speed up the computations
presented in Section \ref{SEC:Experiments} dramatically compared to the naive
approach.

\subsection{Evaluation of Gradients}
\label{SUBSEC:gradient}

To compute the gradient of $J$, we apply the chain rule:  $\nabla J =
(dJ/d\bdf)^T$, where
\[
\ldiff{J}{\bdf} = \lpdiff{L}{\Phi} \ldiff{\Phi}{\hat{\bds}} \ldiff{\hat{\bds}}{\bds} \ldiff{\bds}{\bdu} \ldiff{\bdu}{\bdf}.
\]
Traditionally in PDE-constrained optimization, $dJ/d\bdf$ would be computed
using the \emph{adjoint method} \citep[Section 2.3.3]{Gun2003}, sometimes
called the \emph{reduced gradient method} \citep[Section 12.6]{LY2008}.  In our
context, this method first evaluates
\[
\ldiff{J}{\bdu} = \lpdiff{L}{\Phi} \ldiff{\Phi}{\hat{\bds}} \ldiff{\hat{\bds}}{\bds} \ldiff{\bds}{\bdu}
\]
using the solution $\bdu$ to the PDE corresponding to the value of $\bdf$ at
which $\nabla J(\bdf)$ is desired.  Using the fact that $d\bdu/d\bdf =
\bdA^{-1} \bdB$, it then evaluates
\[
\ldiff{J}{\bdf} = \ldiff{J}{\bdu} \ldiff{\bdu}{\bdf} = \ldiff{J}{\bdu} \bdA^{-1} \bdB
\]
\emph{not} by forming $\bdA^{-1}\bdB$, which is prohibitively expensive, and
then multiplying on the left by $dJ/d\bdu$ but instead by evaluating
$\bdlambda^T = (dJ/d\bdu)\bdA^{-1}$ and then computing $dJ/d\bdf = \bdlambda^T
\bdB$.  The method derives its name from the fact that the equation that must
be solved for $\bdlambda$,
\[
\bdA^T \bdlambda = \left(\ldiff{J}{\bdu}\right)^T,
\]
involves the adjoint of $\bdA$.  Thus, evaluating $\nabla J(\bdf)$ this way
requires two expensive PDE solves:  one with $\bdA$ (the original equation, to
find $\bdu$) and one with $\bdA^T$ (the adjoint equation, to find $\bdlambda$).

By adopting the approach to evaluating $J$ from Section \ref{SUBSEC:shortcut},
we can eliminate \emph{both} of these solves, greatly reducing the cost of
computing $\nabla J$.  For under that approach, by \eqref{EQN:sEqn}, the matrix
\[
\ldiff{\bds}{\bdu} \ldiff{\bdu}{\bdf} = \ldiff{\bds}{\bdf} = \bdY^T \bdB
\]
is already in hand, as $\bdY$ has been pre-computed by solving \eqref{EQN:YEqn}
offline.  We evaluate $dJ/d\bdf$ via
\[
\ldiff{J}{\bdf} = \lpdiff{L}{\Phi} \ldiff{\Phi}{\hat{\bds}} \ldiff{\hat{\bds}}{\bds} \bdY^T \bdB,
\]
using an automatic differentiation package to evaluate $\partial L/\partial
\Phi$, $d\Phi/d\hat{\bds}$, and $d\hat{\bds}/d\bds$, since they involve
differentiating through the neural network $\Phi$.  As our experiments below
show, this approach to evaluating $\nabla J$ is considerably more efficient
than the adjoint method.

\subsection{Computation of Spectrograms}
\label{SSEC:Spectrograms}

We compute the spectrogram $\hat{\bds}$ of the discrete signal $\bds$ in the
usual way via the short-time Fourier transform (STFT) \citep[Section
10.3]{OSB1999}.  We select $L$ frequencies $\omega_0, \ldots, \omega_{L - 1}$,
divide $\bds$ into $M$ overlapping time windows of length $W$, and evaluate
\[
\hat{s}_{\ell, m} = 10 \log_{10} \left| \sum_{k = 0}^{W - 1} w(k)s_{mH + k} e^{-i\frac{2\pi k}{W} \omega_\ell} \right|^2
\]
for $0 \leq \ell \leq L - 1$ and $0 \leq m \leq M - 1$, where $H$ is the
distance between the starting indices of adjacent windows (assumed to be
constant) and $w$ is the von Hann windowing function,
\[
w(k) = \frac{1}{2} - \frac{1}{2}\cos\left(\frac{2\pi k}{W - 1}\right).
\]
The value of $\hat{s}_{\ell, m}$ represents the amount of power present in
$\bds$ at the $\ell$th frequency during the $m$th time window, expressed in dB.

The STFT is a linear operation and therefore can be represented by a matrix
$\bdF$.  If we view the matrix $\hat{\bds}$ as a vector by stacking its
columns, then $\bdF$ has dimensions $LM \times K$ and for each value of $m$ has
a copy of the $L \times W$ matrix
\[
\bdQ = \begin{bmatrix}
w(0) e^{-i\frac{2\pi 0}{W}\omega_0} & \cdots & w(W - 1) e^{-i\frac{2\pi (W - 1)}{W}\omega_0} \\
\vdots & & \vdots \\
w(0) e^{-i\frac{2\pi 0}{W}\omega_{L - 1}} & \cdots & w(W - 1) e^{-i\frac{2\pi (W - 1)}{W}\omega_{L - 1}}
\end{bmatrix}
\]
occupying rows $mL$ through $(m + 1)L - 1$ and columns $mH$ through $mH + W -
1$.  Elsewhere, the entries of $\bdF$ are zero.

\subsection{Enforcing the Frequency Constraint}
\label{SUBSEC:projection}

To restrict the frequency content in $\bdf$, we identify the frequencies among
$\omega_0, \ldots, \omega_L$ that we wish to disallow and construct the
selection matrix $\bdP$ so that $\tilde{\bdF} = \bdP\bdF$ contains exactly the
rows of $\bdF$ corresponding to those frequencies.  (Each row of $\bdP$ is a
row of the $LM \times LM$ identity matrix.)  The constraint $\bdP \bdF \bdf =
0$ then says that $\bdf$ must belong to the nullspace $\sN(\tilde{\bdF})$ of
$\tilde{\bdF}$.

To enforce this constraint during the optimization procedure, we project the
current search direction onto $\sN(\tilde{\bdF})$.  If $\bdf_\nu$ denotes the
guess for the optimal value of $\bdf$ at the $\nu$th optimization step, the
update equation is
\[
\bdf_{\nu + 1} = \bdf_\nu + \alpha \bdPNFtilde \nabla J(\bdf_\nu),
\]
where $\bdPNFtilde$ is the orthogonal projector onto $\sN(\tilde{\bdF})$ and
$\alpha$ is the step size.  We find $\bdPNFtilde$ by noting that since $\bdf$
is real, the constraint $\tilde{\bdF}\bdf = 0$ is equivalent to
\begin{equation}\label{EQN:RealProjConstraint}
\begin{bmatrix}
\Re \tilde{\bdF} \\
\Im \tilde{\bdF}
\end{bmatrix}
\bdf = 0.
\end{equation}
Hence, $\bdPNFtilde$ can be formed by taking the singular value decomposition
of the matrix on the left-hand side of \eqref{EQN:RealProjConstraint} and
extracting the right singular vectors corresponding to zero singular values.

%% file: experiments.tex
We now conduct several experiments that illustrate the effectiveness of our
proposed approach for computing adversarially perturbing signals under a
physical access assumption.

\subsection{Basic Setup}

We take $\Omega$ to be the rectangle $(0, 100) \times (0, 100)$, with the
lengths denominated in meters.  We set $c = \qty{1525}{\meter\per\second}$,
which is a reasonable estimate of the speed of sound in seawater
\citep{KR2007}.  For simplicity, we fix the locations of the submarine and the
interferer in time, setting $x_s(t) = (5, 75)$ and $x_i(t) = (9.75, 68.75)$ for
all $t$.  We position the detector at $x_d = (52.5, 12.5)$.  (Note that while
our framework can accommodate settings in which each of these three agents is
mobile, the method described in Section \ref{SUBSEC:shortcut} for accelerating
the evaluation of the objective requires the location of the detector to be
static.)  We take $\bdP$ to be a bandpass filter for frequencies between
\qty{100}{\hertz} and \qty{6000}{\hertz}.

We use the FEniCS software package to carry out our finite element computations
\citep{FEniCS2015,LMW2012}.  Our mesh for $\Omega$ consists of 23,270
unstructured (but nearly uniformly sized) triangular elements and has 11,836
nodes.  We integrate \eqref{EQN:WaveEqn} to a final time of \qty{1.0}{\second}
using 6000 time steps, yielding a step size of $\Delta t \approx 1.67 \times
10^{-4}$.  For the STFT, we use $L = 129$ uniformly-spaced frequencies between
\qty{0}{\hertz} and \qty{3000}{\hertz} and a window length $W = 64$, which
yields $M = \lceil 6000/64 \rceil = 94$ time windows.  In the context of
\eqref{EQN:DiscreteOptProblem}, these parameters yield decision variables of
the following dimensions: $\bdu \in \mathbb{R}^{\text{71,016,000}}$, $\bdf \in
\mathbb{R}^{6000}$, $\hat{\bds} \in \mathbb{R}^{6000}$, and $\hat{\bds} \in
\mathbb{R}^{129 \times 94}$.

\subsection{Spectrogram Classifiers}

For the detector's classifier $\Phi$, we consider three neural network
architectures from the computer vision literature:  VGG-19
\citep{simonyan2014very}, Inception V3 \citep{szegedy2016rethinking}, and
GoogLeNet \citep{szegedy2015going}. These models are pre-trained for image
classification on the 1.2 million training images and 1000 classes from the
ImageNet Large Scale Visual Recognition Challenge \citep{ILSVRC15}.  To
customize them for spectrogram-based signal detection, we use the standard
technique of \emph{transfer learning} \citep{pmlr-v27-bengio12a}:  we modify
the dimensions of and retrain the networks' final layers specifically for this
task.

We train the classifiers to discriminate between spectrograms generated by pure
sine wave signals $g(t) = \sin(\omega t)$ emitted by potential intruders, where
the frequency $\omega$ is allowed to assume one of the 80 values 10, 20,
\ldots, \qty{800}{\hertz}.  Spectrograms corresponding to signals $g(t)$ with
$\omega \leq \qty{400}{\hertz}$ are labeled malicious (i.e., as coming from our
intruding submarine); those corresponding to higher frequencies are labeled
benign. To add realism, we assume that the signals received by the detector
have been corrupted with noise:  the detector receives the signal $s(t)$ that
results from the signal $g(t)$ propagating through $\Omega$, plus band-limited
noise of the form \citep[Section 14.9.3]{Har2007}
\[
\eta(t) = \sum_{\ell = 0}^{L - 1} A_\ell \cos(\omega_\ell t) + B_\ell \sin(\omega_\ell t),
\]
where $A_\ell$ and $B_\ell$ are independently drawn from a normal distribution
with zero mean and variance $\sigma_\ell^2$, and $\sigma_\ell$ is proportional
to the root-mean-square value of the component of $s(t)$ (the \emph{noiseless}
received signal) belonging to the STFT frequency band $\omega_\ell$.  This way,
the average power of the noise in each band decreases at higher frequencies,
matching the physical reality that higher-frequency signals experience greater
attenuation.

\begin{figure}[t]
\centering
\subfloat[]{\includegraphics[scale = 0.37]{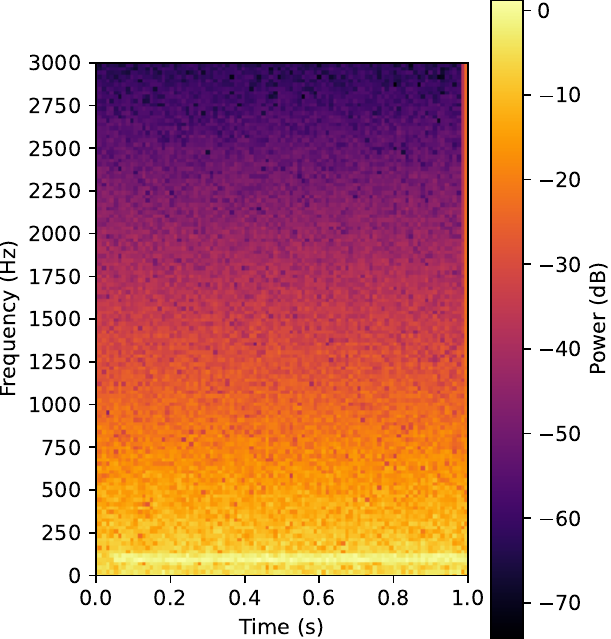}} \quad
\subfloat[]{\includegraphics[scale = 0.37]{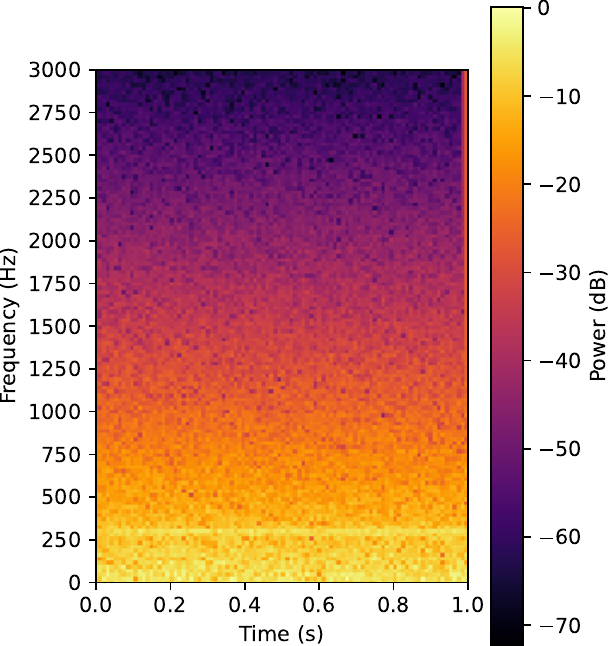}} \quad
\subfloat[]{\includegraphics[scale = 0.37]{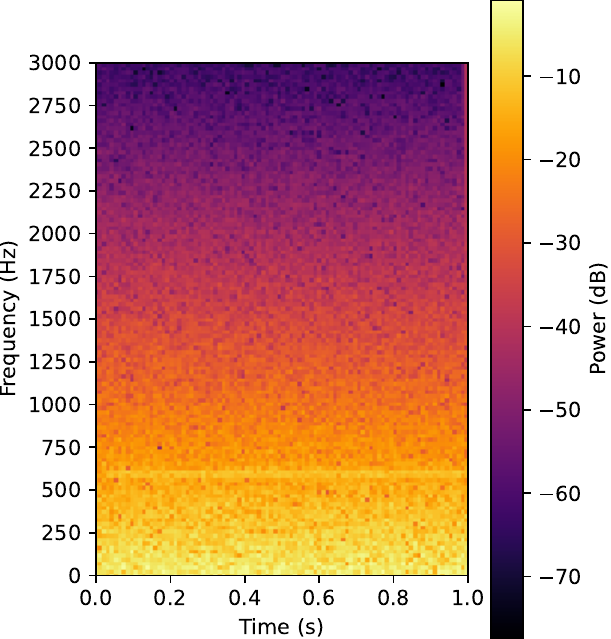}} \quad
\subfloat[]{\includegraphics[scale = 0.37]{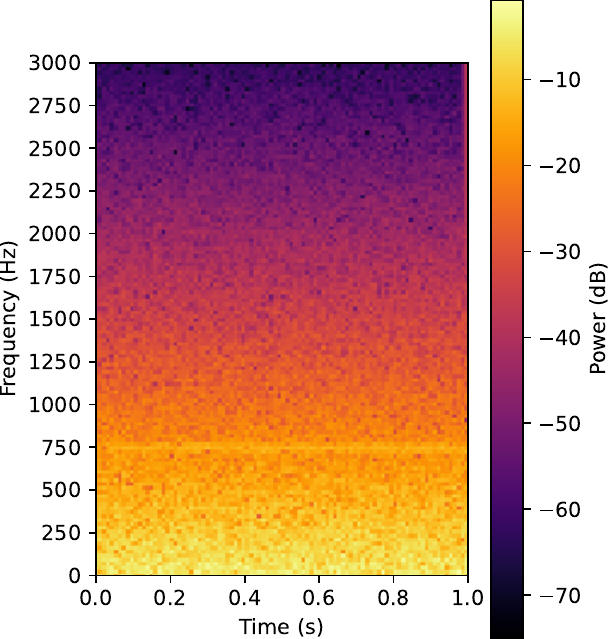}}
\caption{Sample spectrograms from the data used to train the neural networks
$\Phi$ used by the detector. (a) \qty{100}{\hertz} signal. (b)
\qty{300}{\hertz} signal. (c) \qty{600}{\hertz} signal. (d) \qty{750}{\hertz}
signal. Spectrograms containing signals with frequency less than or equal to
\qty{400}{\hertz} are labeled as containing a malicious intruder. Otherwise the
spectrograms are labeled benign.}
\label{FIG:train_specs}
\end{figure}

To generate training data, for each of the 80 allowable values of $\omega$, we
compute the signal $s(t)$ received by the detector in the absence of noise (and
the interferer) by solving the wave propagation problem.  We then generate 40
noisy spectrograms for each frequency by adding $A_\ell + B_\ell i$ to each
entry of the STFT of the corresponding $s(t)$, with $A_\ell$ and $B_\ell$
distributed as just described.  The result is a balanced training set of 3200
observations, examples of which are displayed in Figure \ref{FIG:train_specs}.
We also generate a separate testing data set for monitoring progress during
training and a validation data set for testing our method for computing
perturbing signals, each consisting of 800 spectrograms, 10 for each
value of $\omega$. Finally, because the detector's classification problem is a
binary decision over two classes---malicious or benign---we use a binary
cross-entropy loss both to train the classifiers and as the function $L$ in the
optimization problem \eqref{EQN:OptProblem}.

With the aid of the PyTorch machine learning framework \citep{PyTorch2019}, we
train the neural networks using stochastic gradient descent, terminating the
training when the loss on the testing set stops improving.  The first column of
Table \ref{TAB:Accuracy} shows the accuracy of each model on the spectrograms
in the validation set.  Absent the interferer, all the models are excellent
classifiers, attaining accuracy rates better than 95\%.

\begin{table}[t]
\centering
\begin{tabular}{c c c}
\toprule
Network & \quad Accuracy Without Interferer \quad & \quad Accuracy With Interferer \quad \\
\midrule
Inception V3 & 97.75\% & 0.000\% \\
GoogLeNet    & 97.50\% & 0.000\% \\
VGG-19       & 95.88\% & 0.125\% \\
\bottomrule
\end{tabular}
\caption{Classification accuracy of each network on the validation set.}
\label{TAB:Accuracy}
\end{table}

\subsection{Adversarial Perturbations}
\label{SUBSEC:PerturbationExperiments}

We now test our proposed method for constructing perturbing signals $f(t)$ to
be emitted by the interferer in order to fool the classifiers.  For each
classifier, we take each spectrogram from the validation set that it classified
correctly and solve the discretized optimization problem
\eqref{EQN:DiscreteOptProblem} in search of a perturbing signal that
causes the classifier to misclassify the observation, ideally with high
confidence. Any first-order method for unconstrained problems can
be used to solve \eqref{EQN:DiscreteOptProblem}; we use the limited-memory
BFGS (L-BFGS) method as implemented by \citet{zhu1997algorithm}, computing the
objective and gradient as described in Sections \ref{SUBSEC:shortcut} and
\ref{SUBSEC:gradient} and applying the projected gradient procedure described
in Section \ref{SUBSEC:projection}.  We start the algorithm with a zero initial
guess and use the automatic differentiation capabilities of PyTorch to compute
derivatives involving $\Phi$.

Among perturbing signals which succeed in inducing misclassification, small
signals are preferable to large ones because they require less energy to
emit and are less likely to be detected by a secondary classifier (e.g., a
human observer).  To achieve perturbing signals which are small, we limit L-BFGS to
at most 100 iterations.  Every 10 iterations, we check if misclassification has
been achieved and, if so, terminate the algorithm.

Our results, summarized in the second column of Table \ref{TAB:Accuracy},
demonstrate that misclassification can be achieved consistently.  For both
Inception V3 and GoogLeNet, we are able to induce misclassification of
\emph{every single spectrogram} that they originally classified correctly.  For
VGG-19, we are able to induce misclassification in all but one of the 800
observations in the validation set.  Moreover, empirically, we find that many
of the interfering signals have much smaller amplitude than the signals emitted
by the submarine/intruder. In many cases, the interfering signals are
visually indistinguishable from the background noise.

\begin{figure}[t]
\centering
\subfloat[]{\includegraphics[scale = 0.75]{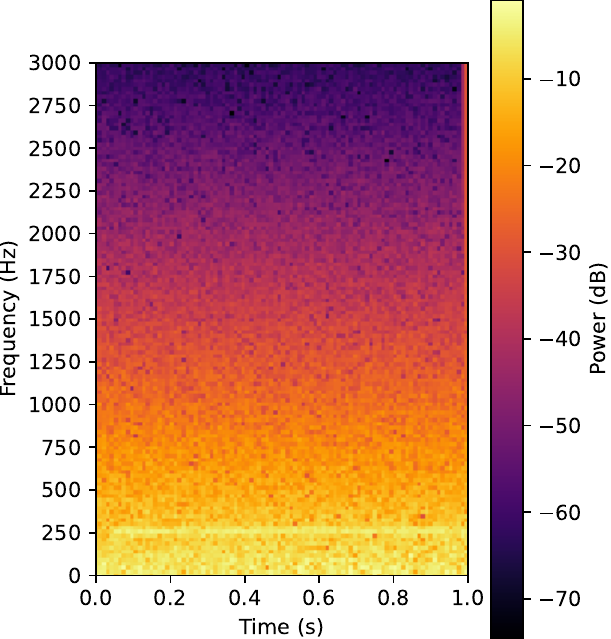}}\quad
\subfloat[]{\includegraphics[scale = 0.75]{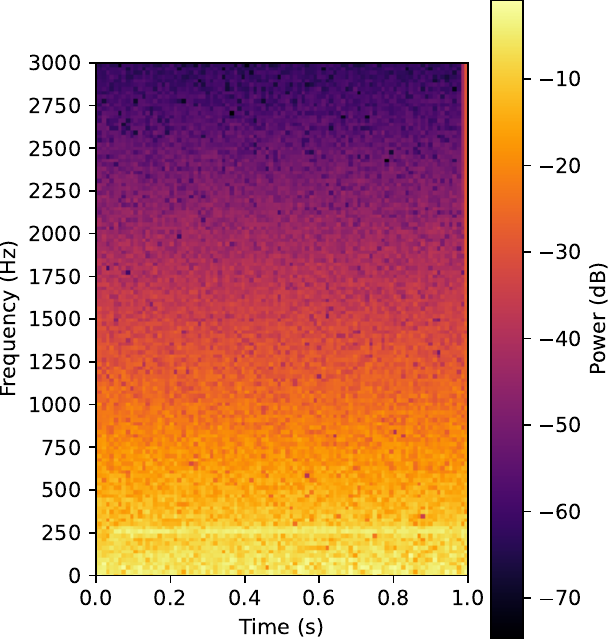}}
\\
\subfloat[]{\label{SFIG:Inv3_example_interferer}\includegraphics[width = \textwidth]{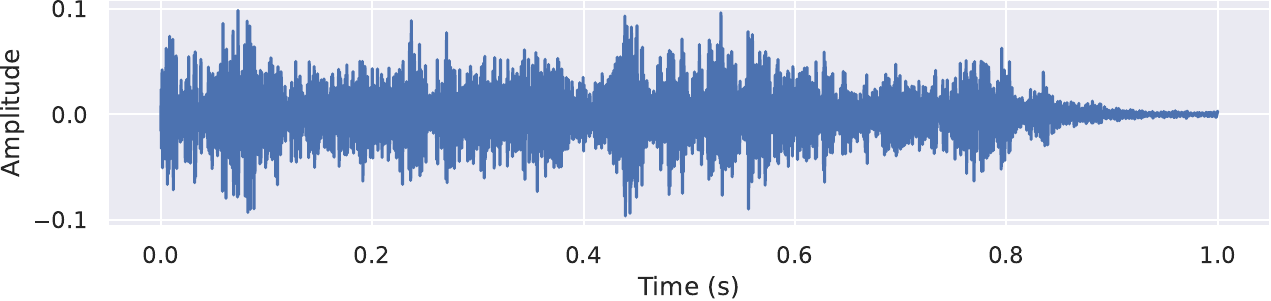}}
\caption{The original (a) was correctly classified as malicious by Inception V3
with 99.0\% confidence. When the interfering signal is added, the perturbed
spectrogram (b) is misclassified by the same network as benign with 99.99\%
confidence. The waveform of the interfering signal is given in (c).}
\label{FIG:Inv3_example}
\end{figure}

Figure \ref{FIG:Inv3_example} provides a typical example of this behavior.  The
figure displays both the original spectrogram and the spectrogram with the perturbing
signal for a \qty{260}{\hertz} intruder signal.  Inception V3 correctly
classifies the original spectrogram as malicious with 99.0\% confidence.  The
interfering signal results in only minor changes to the spectrogram, but the
classifier's confidence that the signal is malicious drops to 0.01\%.

Figure \ref{SFIG:Inv3_example_interferer} depicts the waveform $f(t)$ emitted
by the interferer for this example, which has an amplitude of approximately
0.1.  In these experiments, we always take the intruder signal to have unit
amplitude.  Thus, the interfering signal is about 10 times weaker than the
intruder signal in this case, with the result that it blends easily into the
background noise.

\begin{figure}[t]
\centering
\subfloat[]{\includegraphics[scale = 0.75]{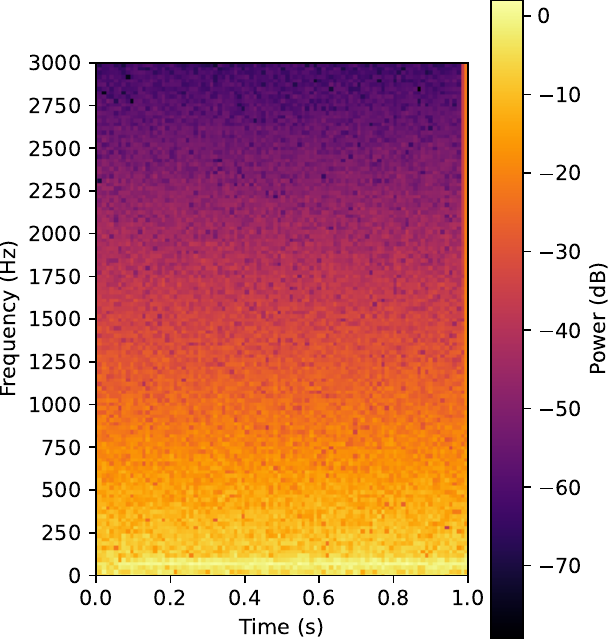}}\quad
\subfloat[]{\includegraphics[scale = 0.75]{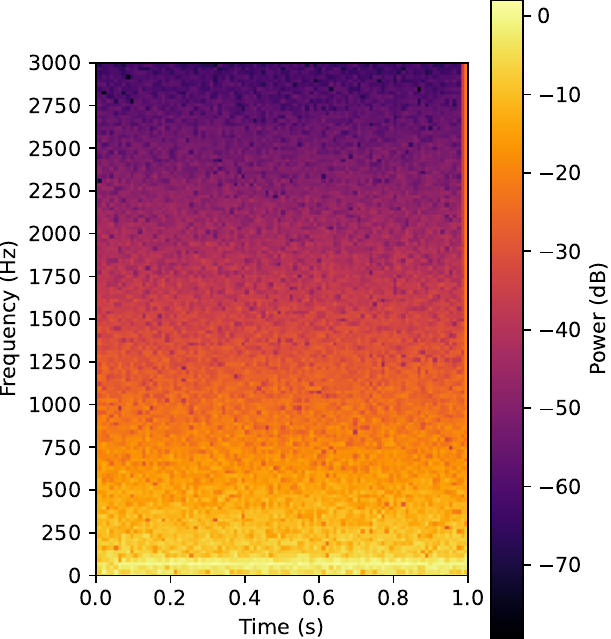}}
\\
\subfloat[]{\includegraphics[width = \textwidth]{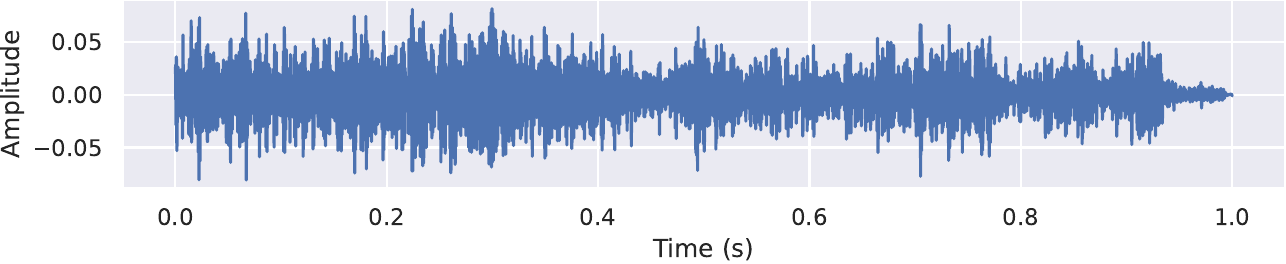}}
\caption{The original (a) was classified as malicious by GoogLeNet with 95.7\%
confidence. When the interfering signal is added, the perturbed spectrogram (b)
is classified by the same network as benign with 99.99\% confidence. Though it
is difficult to find visible differences between these spectrograms, they do
exist.  One example can be found by looking closely in the region corresponding
to \qty{.4}{\second} and \qty{1150}{\hertz}. The waveform of the interfering
signal is given in (c).}
\label{FIG:Googlenet_example}
\end{figure}

Figure \ref{FIG:Googlenet_example} displays a similar example for a
\qty{70}{\hertz} signal initially classified correctly by GoogLeNet.  With the
interfering signal---which is nearly 20 times weaker than the original
signal---in place, the detector's confidence that the signal is malicious drops
from 95.7\% to 0.01\%.

\subsection{Computational Efficiency}

We close with some remarks on the effectiveness of the technique described in
Section \ref{SUBSEC:shortcut} at reducing the time required to complete the
above computations.  Table \ref{TAB:Timings} shows the times (averaged over 800
runs) required to evaluate the objective function and its gradient for each
network architecture using both the naive approach (solving the PDE once for
each objective evaluation, plus an adjoint solve for each gradient evaluation)
and the method of Section \ref{SUBSEC:shortcut}.  These times were computed
using a single node of the ``hamming'' cluster at the High Performance
Computing Center of the Naval Postgraduate School; this node is equipped with
dual 8-core AMD EPYC 7F32 CPUs clocked at 3.7 GHz and 512 GB of RAM.  Note that
in our implementation, the gradient cannot be computed independently of the
objective; hence, the times listed for the gradient computations include the
time required to perform an objective evaluation too.  As the table makes
clear, the method of Section \ref{SUBSEC:shortcut} is substantially faster than
the naive approach by a factor ranging from 60x to 140x.

It is difficult to overstate the significance of these gains.  The PDE problem
we have considered in these proof-of-concept experiments is relatively simple:
the equation is linear and two-dimensional in space; the spatial discretization
is low-order and fairly coarse (only 11,000 degrees of freedom); the temporal
discretization is low-order and explicit; and the horizon for integration in
time is short (only 6,000 time steps).  It takes only a handful of seconds to
solve.  Nevertheless, during optimization, the objective and its gradient may
be evaluated dozens of times, and those seconds quickly add up to minutes---or
even hours---to compute just one perturbing signal if the evaluations
are done the naive way.  Using the approach of Section \ref{SUBSEC:shortcut},
we must solve the PDE (or, rather, its adjoint) only \emph{once} to compute
$\bdY$.  Further evaluations of the objective and gradient then require only
cheap matrix-vector multiplies instead of an expensive solve.  As a result, we
are able to compute all approximately 2400 perturbing signals in Section
\ref{SUBSEC:PerturbationExperiments} with just a few hours of computing time.

With a more physically realistic scenario, each PDE solve may require several
hours of time using specialized high-performance computing hardware.  In such a
case the gains realized by using the method of Section \ref{SUBSEC:shortcut}
would be even more substantial.

\begin{table}[t]
\centering
\begin{tabular}{l l l l}
\toprule
\multirow{2}{*}{Computation} & \multicolumn{3}{c}{Time (s)} \\
& Inception V3 & GoogLeNet & VGG-19 \\
\midrule
Objective (Naive)                         & $6.8 \ttt{0}$  & $6.7 \ttt{0}$  & $6.7 \ttt{0}$  \\
Objective (Section \ref{SUBSEC:shortcut}) & $7.0 \ttt{-2}$ & $4.9 \ttt{-2}$ & $1.0 \ttt{-1}$ \\
Speedup                                   & 97.1           & 136.7          & 67.0           \\
\midrule
Gradient (Adjoint method)                 & $1.3 \ttt{1}$  & $1.3 \ttt{1}$  & $1.3 \ttt{1}$  \\
Gradient (Section \ref{SUBSEC:gradient})  & $1.5 \ttt{-1}$ & $1.1 \ttt{-1}$ & $1.4 \ttt{-1}$ \\
Speedup                                   & 86.7           & 118.2          & 92.9           \\
\bottomrule
\end{tabular}
\caption{Times for evaluating the objective function and its gradient for each
network using the naive/adjoint approaches and the method of Sections
\ref{SUBSEC:shortcut}-\ref{SUBSEC:gradient}.  The speedup factor gained by
using the latter is also listed.}
\label{TAB:Timings}
\end{table}

%% file: conclusion.tex
We have demonstrated that adversarial perturbations can be used to fool neural
network classifiers when the adversary is limited to having only physical
access to the classifiers' inputs.  By considering an example from underwater
acoustics, we have shown how to formulate the problem of computing such
perturbations as a PDE-constrained optimization problem and described how to
solve that problem efficiently.

In addition to being a novel application of PDE-constrained optimization to
machine learning, our results have implications for the real-world use of
neural networks.  Neural networks' lack of robustness, as embodied by their
extreme sensitivity to small changes in their inputs, leaves them ill-suited for
settings in which those inputs can be manipulated by a motivated, sophisticated
adversary and in which misclassification has severe consequences.  Many
applications in security and defense fit this description.  By demonstrating
that indirect manipulation of a network's inputs via a physical environment
suffices to elicit these shortcomings, our work broadens the range of threats
that designers of, e.g., autonomous sensors should consider when choosing
statistical and machine learning models for use in their systems.

This work admits many potentially interesting extensions.  For instance, by
replacing the wave equation with a different PDE (e.g., the Maxwell equations),
one can study the adversarial manipulation of signals whose propagation is
governed by physics other than those of acoustics (e.g., the physics of
electromagnetics).  Another promising direction is the generation of so-called
``universal perturbations'' which can be applied regardless of the background
noise.  Universal perturbations have been successfully constructed in computer
vision under logical access assumptions \citep{MFFF2017}.  Investigating
whether they can also be realized under physical access assumptions is
therefore a natural avenue for future work.